\documentclass[10pt,twocolumn,letterpaper]{article}

\usepackage{cvpr}
\usepackage{times}
\usepackage{epsfig}
\usepackage{graphicx}
\usepackage{amsmath,wrapfig,graphicx}
\usepackage{amssymb,amsthm}
\usepackage{paralist}

\DeclareMathOperator*{\argmin}{arg\,min}
\DeclareMathOperator*{\abs}{abs}
\DeclareMathOperator*{\SO}{\bf SO}
\DeclareMathOperator*{\atan}{atan2}
\DeclareMathOperator*{\logm}{logm}
\DeclareMathOperator*{\expm}{expm}
\DeclareMathOperator*{\relu}{ReLU}
\newtheorem{proposition}{Proposition}

\usepackage{multirow}
\usepackage{colortbl, array}
\usepackage{hhline}

\usepackage[utf8]{inputenc} 
\usepackage[T1]{fontenc}    
\usepackage{hyperref,amsmath,amsthm,amssymb,comment,tabularx} 
\usepackage{url}            
\usepackage{booktabs}       
\usepackage{amsfonts}       
\usepackage{nicefrac}       
\usepackage{microtype}      
\usepackage{authblk}

\usepackage[linesnumbered,ruled]{algorithm2e}
\usepackage{graphicx,picins,caption,paralist,color,wrapfig}
\newtheorem{definition}{Definition}

\newcommand{\highest}[1]{\textcolor{blue}{\mathbf{#1}}}

\definecolor{rulecolor}{RGB}{70,10,171}
\definecolor{tableheadcolor}{RGB}{120,50,200}

\usepackage{algpseudocode}

\graphicspath{{eps/}}

\newcommand{\topline}{ %
        \arrayrulecolor{rulecolor}\specialrule{0.1em}{\abovetopsep}{0pt}}%
\newcommand{\bottomline}{ %
        \arrayrulecolor{rulecolor} \specialrule{\lightrulewidth}{0pt}{0pt}}%

\cvprfinalcopy

\def\cvprPaperID{9}

\ifcvprfinal\fi

\begin{document}

\title{\mbox{SurReal: Fr\'{e}chet Mean and Distance Transform for Complex-Valued Deep Learning}\\[-20pt]}
\author{
Rudrasis Chakraborty, Jiayun Wang, and Stella X. Yu\\
UC Berkeley / ICSI\\
\tt\small\{rudra,peterwg,stellayu\}@berkeley.edu
}
\maketitle

\begin{abstract}
We develop a novel deep learning architecture for naturally complex-valued data, which is often subject to complex scaling ambiguity.
We treat each sample as a field in the space of complex numbers.  With the polar form of a complex-valued number, the general group that acts in this space is the product of planar rotation and non-zero scaling.   This perspective allows us to develop not only a novel convolution operator using weighted Fr\'{e}chet mean (wFM) on a Riemannian manifold, but also a novel fully connected layer operator using the distance to the wFM, with natural equivariant properties to non-zero scaling and planar rotation for the former and invariance properties for the latter.

Compared to the baseline approach of learning real-valued neural network models on the two-channel real-valued representation of  complex-valued data, 
our method achieves surreal performance on two publicly available complex-valued datasets:  MSTAR on SAR images and RadioML on radio frequency signals.  On MSTAR, at $8\%$ of the baseline model size and with fewer than $45,000$ parameters,  our model improves the target classification accuracy from $94\%$ to $98\%$ on this highly imbalanced dataset.  On RadioML, our model achieves comparable RF  modulation classification accuracy at $10\%$ of the baseline model size.
\end{abstract}

\section{Introduction}\label{sec1}

We study the task of extending deep learning to naturally complex-valued data, where useful information is intertwined in both magnitudes and phases.  For example, synthetic aperture radar (SAR) images, magnetic resonance (MR) images,  and radio frequency (RF) signals are acquired in complex numbers, with the magnitude often encoding the amount of energy and the phase indicating the size of contrast or geometrical shapes.  Even for real-valued images, their complex-valued representations could be more successful for many pattern recognition tasks; the most notable examples are the Fourier spectrum and spectrum-based computer vision techniques ranging from steerable filters \cite{freeman1991design} to spectral graph embedding \cite{maire2016affinity,yu2012angular}.

A straightforward solution is to treat the complex-valued data as two-channel real-valued data and apply real-valued deep learning.  Such an Euclidean space embedding would not respect the intrinsic geometry of complex-valued data.  For example, in MR and SAR images, the pixel intensity value could be subject to  complex-valued scaling.  One way to get around such an ambiguity is to train a model with data augmentation \cite{Krizhevsky2012,Dieleman2015,wang2017joint}, but such extrinsic data manipulation is time-consuming and ineffective.  Ideally, deep learning on such images should be invariant to the group of non-zero scaling and planar rotation in the complex plane.  

We treat each complex-valued data sample as a field in the space of complex numbers, which is a special non-Euclidean space.  This perspective allows us to develop novel concepts for both convolution and fully connected layer functions that achieve equivariance and invariance to complex-valued scaling.

A major hurdle in extending convolution from the Euclidean space to a non-Euclidean space is the lack of a vector space structure.  In the Euclidean space, there exists a translation to go from one point to another, and convolution is equivariant to translation.  In a non-Euclidean space such as a sphere, a point undergoing translation may no longer remain in that space, hence translation equivariance is no longer meaningful.  What is essential and common between a non-Euclidean space and the Euclidean space is that, there is a group that transitively acts in the space.  For example, there is a rotation, {\it instead of translation}, to go from one point to another on a sphere.  Extending convolution to a non-Euclidean space should consider equivariance to some transitive action group specific to that space.  

Note that such a manifold view applies to both the domain and the range of the data space.  To extend deep learning to complex-valued images or signals, we take the manifold perspective towards the {\it range} space of the data.  

There is a long line of works that define convolution in a non-Euclidean space by treating each data sample as a function in that space \cite{worrall2017harmonic,cohen2016group,cohen2018spherical,esteves2017polar,chakraborty2018h,kondor2018generalization}. 

Our key insight is to represent a complex number by its polar form, such that the general group that acts in this space is the product of planar rotation and non-zero scaling.    This representation turns the complex plane into a particular Riemannian manifold.
 We want to define convolution that is equivariant to the action of this product group in that space.
 
When a sample is a field on a Riemannian manifold, 
\begin{itemize}
\setlength{\itemsep}{0mm}
\item  Convolution defined by weighted Fr\'{e}chet mean (wFM) \cite{Frechet1948a}  is equivariant to the group that naturally acts on that manifold \cite{chakraborty2018manifoldnet}.
\item  Non-linear activation functions such as ReLU may not be needed, since wFM is a non-linear contraction mapping \cite{Mallat2016} analogous to ReLU or sigmoid.
\item Taking the Riemannian geometric point of view, we could also use tangent ReLU for better accuracy.
\item We further propose a distance transform as a fully-connected layer operator that is invariant to complex scaling.   It takes complex-valued responses at a previous layer to the real domain, where all kinds of standard CNN functions can be subsequently used.
\end{itemize}

A neural network equipped with our wFM filtering and distance transform on complex-valued data has a group invariant property similar to the standard CNN on real-valued data.  Existing complex-valued CNNs tend to extend the real-valued counterpart to the complex domain based on the form of functions \cite{bunte2012adaptive,trabelsi2017deep}, e.g. convolution or batch normalization.
None of complex-valued CNNs are derived by studying the desired property of functions, such as equivariance or linearity.   Our complex-valued CNN is composed of layer functions with all the desired properties and is a theoretically justified analog of the real-valued CNN.

On the SAR image dataset MSTAR, compared to the baseline of a real-valued CNN acting on the two-channel real representation of complex-valued data and reaching $94\%$ accuracy,
our complex-valued CNN acting directly on the complex-valued data (i.e., also without any preprocessing) achieves $98\%$ target classification accuracy with only $8\%$ of  parameters.  Likewise,
on the radio frequency signal dataset RadioML, our method achieves comparable modulation mode classification (a harder task than target recognition) performance with fewer parameters.

To summarize, we make two major contributions.
\begin{enumerate}
\setlength{\itemsep}{0mm}
\item We propose novel complex-valued CNNs with theoretically proven  equivariance and invariance properties.
\item We provide {\it sur-real} (pun intended) experimental validation of our method on complex-valued data classification tasks, demonstrating significant performance gain at a fraction of the baseline model size.
\end{enumerate}
These results demonstrate significant benefits of designing new CNN layer functions with desirable properties on the complex plane as opposed to applying the standard CNN to the 2D Euclidean embedding of complex numbers.

\section{Our Complex-Valued CNN Theory}\label{theory}

We first present the geometry of the manifold of complex numbers and then
develop complex-valued convolutional neural network (CNN) on that manifold.  

\noindent
{\bf Space of complex numbers.}  Let $\mathbf{R}$  denote the set of real numbers.  All the complex number elements assume the form $a+ib$, where $i=\sqrt{-1}$, $a,b \in \mathbf{R}$, and lie on a  a Riemannian manifold \cite{boothby1986introduction} denoted by $\mathbf{C}$.  The distance induced by the canonical Riemannian metric is:
\begin{align}
\label{theory:eq0}
d(a+ib, c+id) = \sqrt{(a-c)^2+(b-d)^2}.
\end{align}

We identify $\mathbf{C}$ with the polar form of complex numbers.
\begin{definition}
\label{def1}
We  identify each complex number, $a+ib$, with its polar form, $r\exp(i\theta)$, where $r$ and $\theta$ are the {\it absolute value ($\abs$) or magnitude}  and {\it argument ($\arg$) or phase}  of $a+ib$. Here $\theta \in [-\pi, \pi]$.  Hence, we can identify $\mathbf{C}$ as $\mathbf{R}^+ \times \SO(2)$, where
$\mathbf{R}^+$ is the set of positive numbers, and $\SO(2)$ is the manifold of planar rotations. Let $F: \mathbf{C}\rightarrow \mathbf{R}^+ \times \SO(2)$ be the mapping from the complex plane to the  product manifold $\mathbf{R}^+ \times \SO(2)$:
\begin{align*}
a+ib & \overset{F}{\mapsto} \left(r, R(\theta)\right),\\
r &=\abs(a+ib)= \sqrt{a^2+b^2}\\
\theta &= \arg(a+ib)=\atan(b,a)\\
R(\theta) &= \begin{bmatrix} \cos(\theta) & -\sin(\theta) \\ \sin(\theta) & \cos(\theta)\end{bmatrix}.
\end{align*}
\end{definition}
\noindent
Note that $F$ is bijective. 

\noindent
{\bf Manifold distance between complex numbers.}
The geodesic distance on this manifold is the Euclidean distance induced from Eq. \eqref{theory:eq0} in the tangent space. Given $\mathbf{z}_1, \mathbf{z}_2 \in \mathbf{C}$, let $(r_1, R_1) = F(\mathbf{z}_1)$ and $(r_2, R_2) = F(\mathbf{z}_2)$.  While the Euclidean distance between two complex numbers is Eq. \eqref{theory:eq0}, their manifold distance $\mathbf{R}^+ \times \SO(2)$ is:
\begin{align}
\label{theory:eq1}
d\left(\mathbf{z}_1, \mathbf{z}_2\right) = \sqrt{\log^2(r_1^{-1}r_2)+\|\logm\left(R_1^{-1}R_2\right)\|_F^2}, 
\end{align} 
where $\logm$ is the matrix logarithm. Note that, for $A =R(\theta) \in \SO(2)$, we choose $\logm(A)$ to be $\theta \begin{bmatrix} 0 & 1 \\ -1 & 0\end{bmatrix}$.

\noindent
{\bf Transitive actions and isometries.}
$\mathbf{C}$ is in fact a {\it homogenous Riemannian manifold} \cite{helgason1962differential}, a topological space on which there is a group of actions acts transitively \cite{dummit2004abstract}.

\begin{definition}
\label{def2}
Given a (Riemannian) manifold $\mathcal{M}$ and a group $G$, we say that $G$ acts on $\mathcal{M}$ (from left) if there exists a  mapping $L: \mathcal{M} \times G \rightarrow \mathcal{M}$ given by $\left(X, g\right) \mapsto g.X$ satisfies \begin{inparaenum}[\bfseries (a)] \item $L\left(X, e\right) = e.X = X$ \item $(gh).X = g.(h.X)$ \end{inparaenum}. 
 An action is called a transitive action {\it iff} given $X, Y \in \mathcal{M}$, $\exists g \in G$, such that $Y = g.X$.  
\end{definition}

\begin{proposition}
\label{theory:prop1} Group $G:=\left\{\mathbf{R}\setminus \{0\}\right\} \times \SO(2)$ transitively acts on $\mathbf{C}$ and the action is given by $\left(\left(r, R\right), \left(r_g, R_g\right)\right) \mapsto \left(r_g^2r, R_gR\right)$.
\end{proposition}
It is straightforward to verify that group $G$ transitively acts on $\mathbf{C}$.  We show that $G$ is the set of isometries on $\mathbf{C}$.

\begin{proposition}
\label{theory:prop2}
Given $\mathbf{z}_1 = (r_1, R_1), \mathbf{z}_2 = (r_2, R_2) \in \mathbf{C}$ and $g = (r_g, R_g) \in G$, $d\left(g.\mathbf{z}_1, g.\mathbf{z}_2\right) = d\left(\mathbf{z}_1, \mathbf{z}_2\right)$.
\end{proposition}
\noindent 
The proof follows from the definitions of $d$ and $g$:
\begin{align*}
&d\left(g.\mathbf{z}_1, g.\mathbf{z}_2\right) \\ =&\sqrt{\log^2\left((r_g^2r_1)^{-1}(r_g^2r_2)\right)+  
\|\logm\left(R_1^{-1}R_g^{-1}R_gR_2\right)\|_F^2} \\
=&d\left(\mathbf{z}_1, \mathbf{z}_2\right).  \hspace{2.3in} \qed
\end{align*}

Having defined our manifold range space for complex numbers, we focus on extending two key properties, {\it equivariance} of a convolution operator and  {\it invariance} of a CNN, from real-valued CNNs to complex-valued CNNs.

\noindent
{\bf Equivariance property of convolution.}  In the Euclidean space $\mathbf{R}^n$, the convolution operator is equivariant to translation: Given the kernel of convolution, if the input is translated by $\mathbf{t}$, the output would also be translated by $\mathbf{t}$.  This property enables weight sharing across the entire spatial domain of an image.  The group of translations is the group of isometries for $\mathbf{R}^n$, and  it transitively acts on $\mathbf{R}^n$.  

We extend these concepts to our complex number manifold $\mathbf{C}$. Our 
$G = \left\{\mathbf{R}\setminus \{0\}\right\} \times \SO(2)$ transitively acts on $\mathbf{C}$ and is the group of isometries.  In order to generalize the Euclidean convolution operator on $\mathbf{C}$, we need to define an operator on $\mathbf{C}$ which is equivariant to the action of $G$. 

CNNs on manifold valued data have recently been explored in \cite{chakraborty2018manifoldnet}, where convolution is defined 
on manifold $\mathcal{M}$ and equivariant to the group $G$ that acts on $\mathcal{M}$.  In our case, manifold $\mathcal{M}\! =\! \mathbf{C}$ and action group $G =  \{\mathbf{R}\!\setminus\! \{0\}\} \!\times\! \SO(2)$.  

\noindent
{\bf Convolution as manifold Fr{\'e}chet mean filtering.} 
Given $K$ points on our manifold $\mathbf{C}$:  $\left\{\mathbf{z}_i\right\}_{i=1}^K\!\subset\! \mathbf{C}$, and $K$ nonnegative weights  $\left\{w_i\right\}_{i=1}^K \!\subset\! (0,1]$ with $\sum_i w_i=1$, the  {\it weighted Fr{\'e}chet mean (FM)} ($\textsf{wFM}$) is defined as \cite{Frechet1948a}:
\begin{align}
\label{theory:eq2}
\textsf{wFM}\left(\left\{\mathbf{z}_i\right\}, \left\{w_i\right\}\right) = \argmin_{\mathbf{m}\in \mathbf{C}} \sum_{i=1}^K w_id^2\left(\mathbf{z}_i, \mathbf{m}\right),
\end{align}
where $d$ is the distance defined in Eq. \eqref{theory:eq1}.   Unlike the standard Euclidean convolution which {\it evaluates} the weighted data mean given the filter weights, the manifold convolution 
wFM {\it solves} the data mean that minimizes the weighted variance.  There is no closed-form solution to wFM; however, there is a provably convergent $K-$step iterative solution \cite{chakraborty2018manifoldnet}. 

While our filter response 
$\textsf{wFM}\left(\left\{\mathbf{z}_i\right\}, \left\{w_i\right\}\right) \in \mathbf{C}$ is complex-valued, a minimizing argument to Eq. \eqref{theory:eq2}, the filter weights $\left\{w_i\right\}$ themselves are real-valued.   They are learned through stochastic gradient descent, subject to additional normalization and convexity constraints on $\left\{w_i\right\}$.  

\begin{figure}[!ht]
\includegraphics[width=0.36\textwidth]{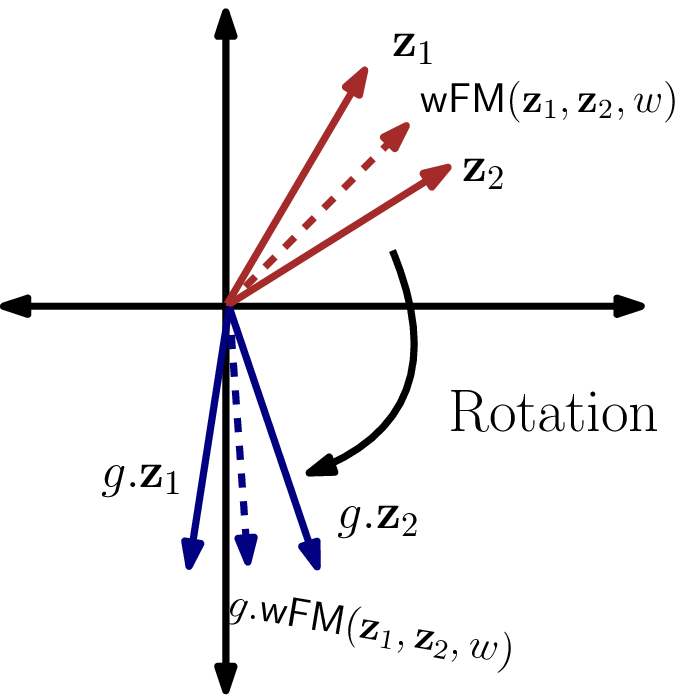}\\
\includegraphics[width=0.49\textwidth]{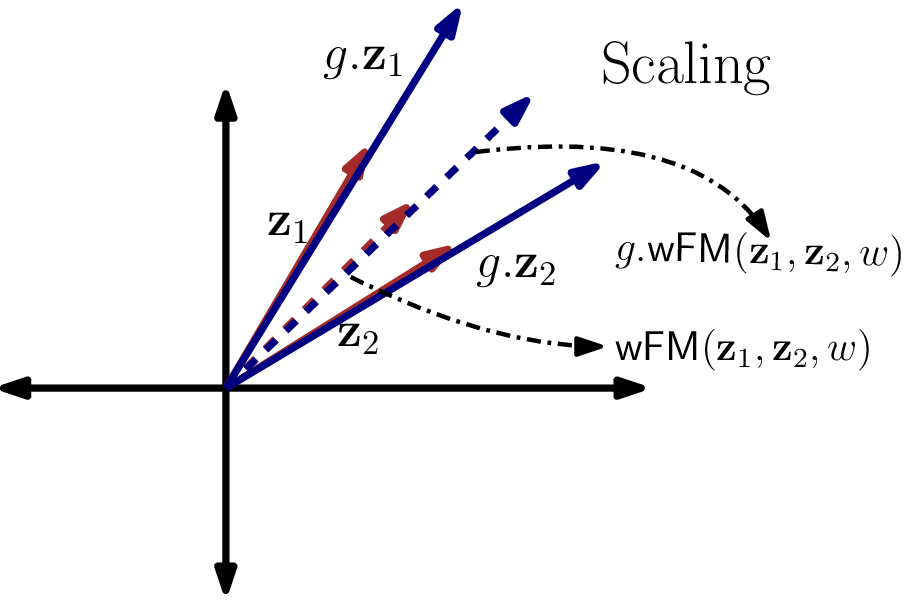}
\caption{Equivariance of weighted Fr{\'e}chet mean filtering with respect to rotation and scaling in the complex plane.}\label{fig:fig0}
\end{figure}

\begin{proposition}
\label{theory:prop3}
The convolution definition in Eq. \eqref{theory:eq2} is equivariant to the action of $G = \left\{\mathbf{R}\setminus \{0\}\right\} \times \SO(2)$.
\end{proposition}
The equivariance property of convolution follows from the isometry in Prop. \eqref{theory:prop2}.
Fig \eqref{fig:fig0} illustrates the equivariance of wFM with respect to planar rotation and scaling. 

\noindent
{\bf Manifold vs. Euclidean convolution.}  Convolution is often written as $\sum_i w_i x_i$, where $\left\{w_i\right\}$ is the filter and $\left\{x_i\right\}$ is the signal.  With our convexity constraint on $\left\{w_i\right\}$, $\sum_i w_i x_i$ is the wFM on the Euclidean space as it is the minimizer of the weighted variance defined in Eq. \eqref{theory:eq2}.  The convexity constraint is to ensure that the resultant stays on the manifold.  Therefore, wFM as a convolution operator on the manifold might appear rather arbitrary at first glance, it is an obvious choice if we regard the standard convolution as the minimizer of the weighted variance in the Euclidean space.

Next we turn to nonlinear activation functions.  Our wFM is non-linear and {\it contractive} \cite{chakraborty2018manifoldnet}, it thus performs not only convolution but also nonlinear activation to a certain extent.  Nevertheless, we extend ReLU in the Euclidean space to a manifold in a principled manner.

\noindent
{\bf ReLU on the manifold: tReLU.}  The tangent space of a manifold could be regarded as a local Euclidean approximation of the manifold, and a pair of transformations, logarithmic and exponential maps, establish the correspondence between the manifold and the tangent space.

Our tReLU is a function from $\mathbf{C}$ to $\mathbf{C}$, just like the Euclidean ReLU  from $\mathbf{R}^n$ to $\mathbf{R}^n$, but it is composed of three steps: {\bf 1)} Apply logarithmic maps to go from a point in $\mathbf{C}$ to a point in its tangent space; 
{\bf 2)} Apply the Euclidean ReLU in the tangent space; {\bf 3)}
Apply exponential maps to come back to $\mathbf{C}$ from the tangent space.
\begin{align}
&(r, R) \overset{\text{tReLU}}{\mapsto} \nonumber\\
&\left(\; \exp(\relu(\log(r))),
\expm(\relu(\logm(R))) \;\right) 
\end{align}
where $\expm$ is the matrix exponential operator.
Our manifold perspective leads to a non-trivial extension of ReLU, partitioning the complex plane by $r$ and $\theta$ into four scenarios, e.g., those with $r\!<\!1$ would be rectified to $r\!=\!1$.

\noindent
{\bf Invariance property of a CNN classifier.} 
For classification tasks, having equivariance of convolution and range compression of nonlinear activation functions are not enough; we need the final representation of a CNN invariant to within-class feature variations.  

In a standard Euclidean CNN classifier, the entire network is invariant to the action of translations, achieved by the fully connected (FC) layer.  Likewise, we develop a FC  function on $\mathbf{C}$ that is invariant to the action of $G$. 

\noindent
{\bf Distance transform as an invariant FC layer.}  Since our distance $d$ is shown invariant to $G$,
we propose the distance of each point in a set to their weighted Fr{\'e}chet mean, which is equivariant to $G$, as a new FC function on $\mathbf{C}$.  

Consider turning an $m$-channel $s$-dimensional feature representation,  $\left\{\mathbf{t}_i\right\}_{i=1}^m \!\subset\! \mathbf{C}$, into a single FC feature $u$ of $m$ dimensions.  Each input channel $\mathbf{t}_i$ contains $s$ elements (in any matrix shape) and is treated as an $s$-dimensional feature vector. Our distance transform first computes the wFM of $m$ input features and then turns input channel $i$ into a single scalar $u_{i}$ as its distance to the mean:
\begin{align}
\mathbf{m} &= \textsf{wFM}(\{\mathbf{t}_i\}, \{v_i\})\\
u_{i} &= d(\mathbf{t}_i, \mathbf{m}),
\end{align}
The $m$ filter weights $v_i$ are learned per FC output channel, and there could be multiple output channels in the FC layer.  

\begin{proposition}
\label{theory:prop4}
The above distance transform, defined as the distance to the wFM, is invariant to the action of $G$.
\end{proposition}
\noindent
The proof follows from Propositions \ref{theory:prop2} and \ref{theory:prop3}:
\begin{align*}
& d(g.\mathbf{t}_i, \textsf{wFM}(g.\{\mathbf{t}_i\}, \{v_i\}))&\\
=&d(g.\mathbf{t}_i, g. \textsf{wFM}(\{\mathbf{t}_i\}, \{v_i\})) 
&\text{equivariance of wFM}\phantom{.\qed}\\
=&d(\mathbf{t}_i, \textsf{wFM}(\{\mathbf{t}_i\}, \{v_i\}))
&\text{invariance of distance}.\qed
\end{align*}

With our distance transform,  complex-valued intermediate feature representations are turned into real values, upon which we can apply any of the standard layer functions in the real domain, such as softmax to the last layer of $c$ channels for $c$-way classification.  

\begin{figure}[htp]
\centering
\includegraphics[width=0.49\textwidth]{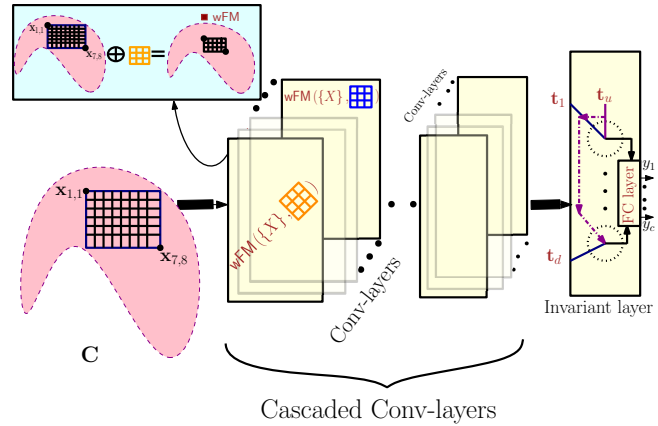}
\caption{Sample architecture of our complex-valued CNN classifier that is invariant to $G$.  It consists of our newly proposed wFM convolution, tReLU nonlinear activation, and FC distance transform layer functions, together achieving invariance to complex-valued scaling in the range space.
}\label{fig:fig1}
\end{figure}

\noindent
{\bf Complex-valued neural network.} With these new convolution, nonlinear activation, and FC layer functions, we can construct a complex-valued CNN which is invariant to the action of $G$. 
Fig. \eqref{fig:fig1} illustrates a possible CNN architecture.
Alg. \eqref{alg:complexnet} presents a CNN work-flow with two convolution layers and one FC layer.

{\begin{algorithm}[]
 		\caption{ \label{alg:complexnet} Workflow of our sample CNN classifier with 2 convolution layers and 1 FC layer.}
 		\begin{algorithmic}
 		    \Function {Complex-Valued CNN variables}{$c^1_{in}, c^1_{out}, k_1, c^2_{out}, k_2, l, c$}
 		        \State $x \leftarrow Input(c^1_{in}, h, w)$
 		        \State $x \leftarrow Conv(x, c^1_{out}, k_1)$
 		        \State $x \leftarrow tReLU(x)$
 		        \State $x \leftarrow Conv(x, c^2_{out}, k_2)$
 		        \State $x \leftarrow tReLU(x)$
 		        \State $x \leftarrow Inv(x, l, c)$
 			\EndFunction
 		\end{algorithmic}
\end{algorithm}}

\section{Experimental Results}\label{sec3}
We conduct our experiments on two publicly available complex-valued datasets: MSTAR \cite{keydel1996mstar} and RadioML \cite{convnetmodrec,rml_datasets}.  MSTAR contains complex-valued 2D SAR images, and RadioML contains complex-valued 1D RF signals. 

\subsection{MSTAR Experiments}

\begin{figure}[!hbp]
\centering
\includegraphics[width=0.49\textwidth,clip]{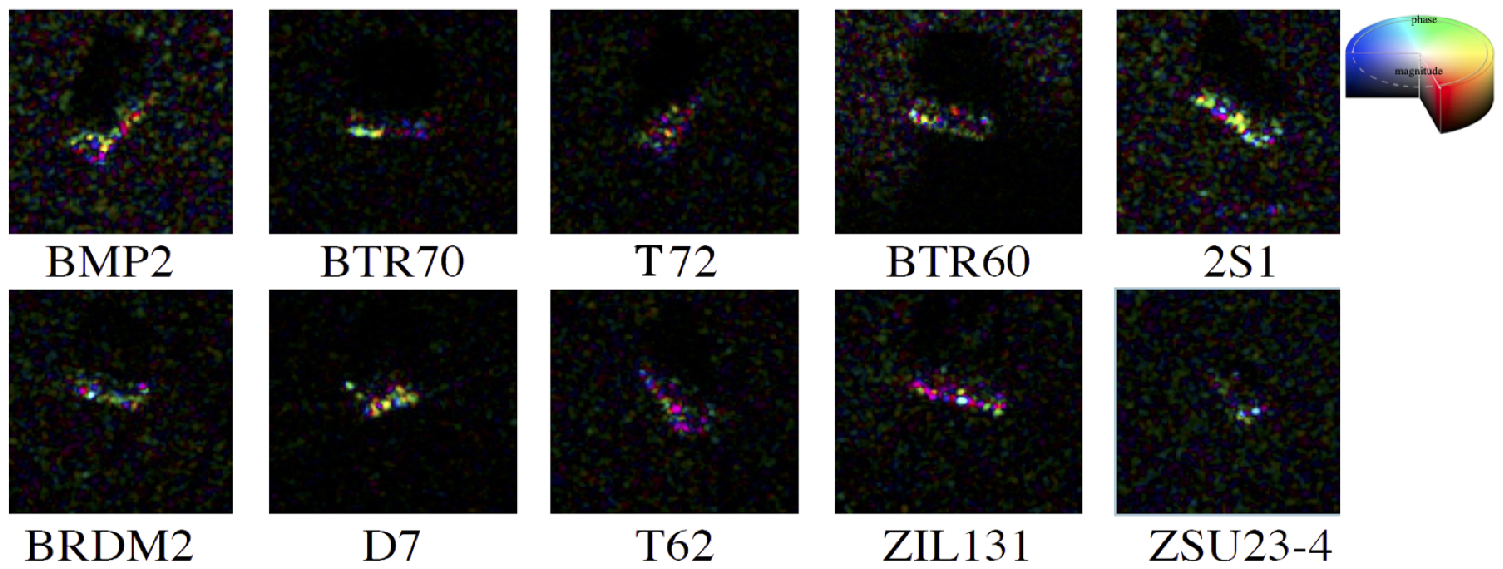}\\
\includegraphics[width=0.49\textwidth,clip]{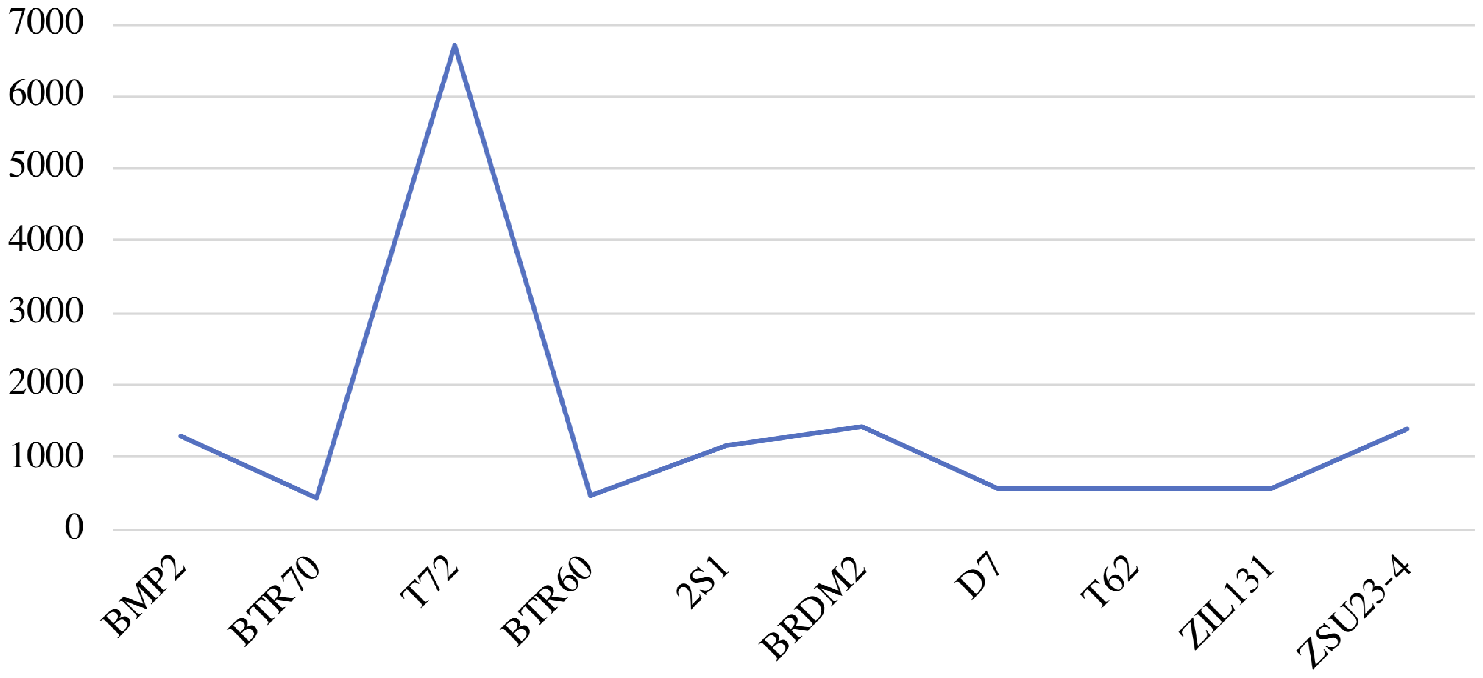}\vspace{-3mm}
\caption{MSTAR has 10 imbalanced target classes, a sample image per class shown on the top and the total number of images per class shown in the bottom. We use the HSV color map for visualizing the phase of a complex-valued image.  SAR images are noisy with a large intensity range.}\label{fig0}
\end{figure}

\noindent
{\bf MSTAR dataset.} It consists of X-band SAR image chips with 0.3m $\times$ 0.3m resolution of $10$ target classes such as infantry combat vehicle (BMP2) and armored personnel carrier BTR70.   The number of instances per class varies greatly from $429$ to $6694$.  We crop $100\times 100$ center regions from each image without other preprocessing (Fig. \eqref{fig0}).

\noindent
{\bf MSTAR baselines.}  We use the real-valued CNN model in Fig. \eqref{fig1} and consider 4 possible representations of
complex-valued  inputs as real-valued data.    Let $\mathbf{z}=a+ib=r e^{j\theta}$.
\begin{enumerate}
\setlength{\itemsep}{-1mm}
\item $(a,b)$: Treat a 1-channel complex-valued image as a  2-channel real-valued image, with real and imaginary components in two separate channels.
\item $r$: Take only the absolute value of a complex-valued image to make a 1-channel real-valued image, with the phase of complex numbers ignored.
\item $(a,b,r)$: Take both the real, imaginary, and magnitude of a complex-valued image to make a 3-channel real-valued image.  
\item $(r,\theta)$: Take the magnitude and phase of a complex-valued image to make a 2-channel real-valued image. 
\end{enumerate}
We perform a $30$-$70$ random train-test split and report the average classification accuracy over $10$ runs.

\begin{figure}[!tp]
  \centering
\begin{tabular}{c}  
\includegraphics[width=0.49\textwidth,clip]{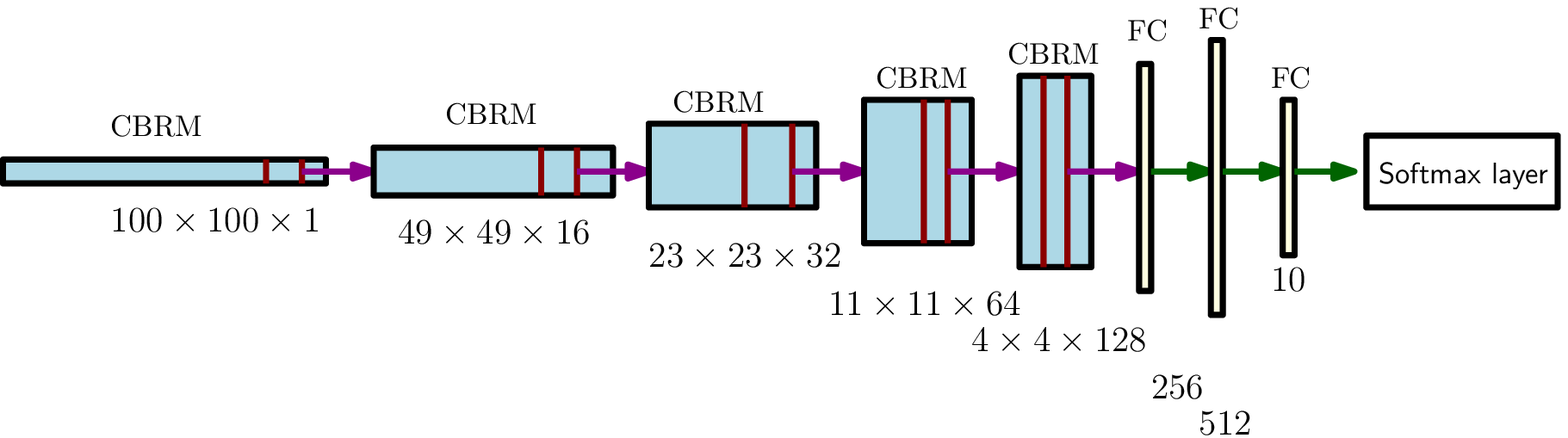}\\
\includegraphics[width=0.49\textwidth,clip]{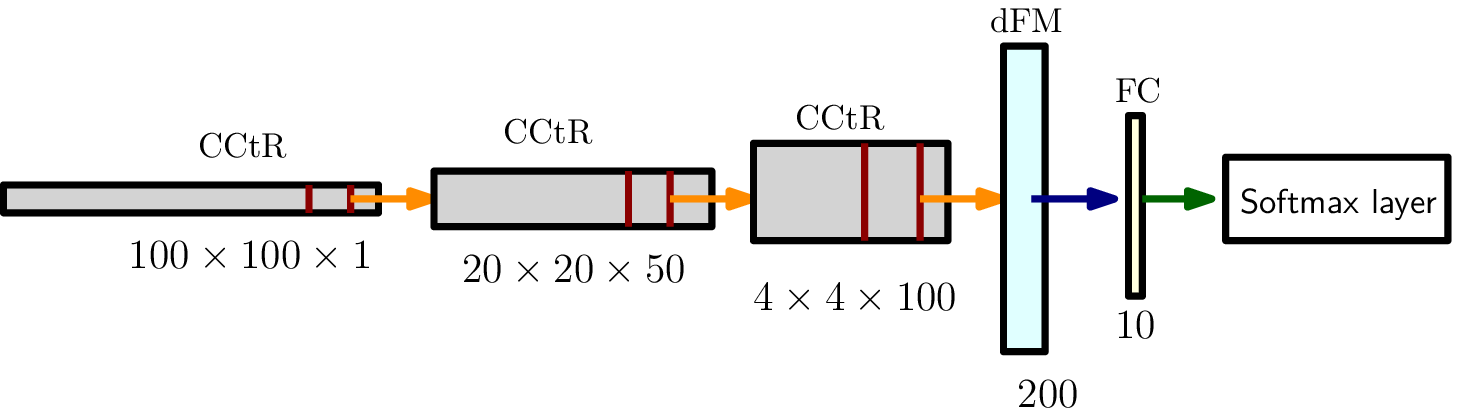}
\end{tabular}
\caption{Real-valued CNN  baseline model (top) and our complex-valued model (bottom) for MSTAR.  CBRM denotes Conv, Batch-Normalization, ReLU,  Pooling. CCtR denotes Complex Convolution, tangent ReLU. 
}\label{fig1}
\end{figure}

\noindent
{\bf Our CNN model.}  We use two complex convolution layers with kernel size $5\times 5$ and stride $5$ followed by one complex convolution layer with kernel size $4\times 4$ and stride $4$, then we use an invariant last layer with a softmax layer at the end for classification. For the three complex convolution layers, the number of output channels are $50$, $100$ and $200$ respectively. We use ADAM optimizer with learning rate $0.005$ and mini-batch size $100$. 

\noindent
{\bf MSTAR results.}
Table \eqref{tab1} shows the
 confusion matrix and the overall classification accuracy for each of the four real-valued CNN baseline and our complex-valued CNN model.
Ours has a $3.6\%$ accuracy gain over the best baseline.  

\begin{table}
\centering
\scalebox{0.72}{
\begin{tabular}{@{}|c|c|c|c|c|c|c|c|c|c|@{}}

\multicolumn{3}{c}{$\textcolor{red}{(a,b): 89.77\%}$}\\
\topline
$\highest{84.5}$  & 2.1   & 0.9    & 11.7  &    & 0.6   &     & 0.2   & 0.1   &    \\
0.2   & $\highest{78.3} $ &     & 21.2  &    &    &     & 0.2   &    &    \\
0.5   &    & $\highest{94.2}$   & 0.9   & 0.2   & 0.1   &     & 3.8   &    & 0.2   \\
   & 0.7   &     & $\highest{99.3}$  &    &    &     &    &    &    \\
0.8   & 1.6   & 0.4    & 4.6   & $\highest{81.7}$  & 6.2   &     & 4.6   & 0.1   & 0.1   \\  \midrule
0.1   &    &     & 5.3   & 0.1   & $\highest{94.1}$  &     &    & 0.4   &    \\
   &    & 4.2    & 0.3   &    & 1.2   & $\highest{88.5}$   & 2.1   & 1.9   & 1.7   \\
   &    & 7.7    & 4.4   & 0.2   & 0.2   &     & $\highest{87.6}$  &    &    \\
   &    & 4.2    & 1.2   & 0.5   & 0.5   &     & 0.5   & $\highest{93.0}$  &    \\
0.1   &    & 8.9    & 2.4   &    & 8.2   & 0.6    & 3.1   & 0.4   & $\highest{76.4}$  \\
\bottomline

\multicolumn{3}{c}{ $\textcolor{red}{r: 94.46\%}$}\\
\topline
$\highest{95.3}$  & 4.0   & 0.5    &    & 0.2   &    &     &    &    &    \\
   & $\highest{98.6}$  & 0.7    &    & 0.7   &    &     &    &    &    \\
0.4   & 0.1   & $\highest{99.2 }$  &    &    & 0.1   &     & 0.1   &    & 0.1   \\
0.9   & 65.4  & 4.7    & $\highest{22.2}$  & 1.8   & 0.4   &     & 4.7   &    &    \\
0.1   & 3.4   & 1.1    &    & $\highest{94.0}$  & 1.0   &     & 0.1   &    & 0.3   \\\midrule
2.9   & 0.6   & 0.3    &    & 0.4   & $\highest{94.4}$  & 0.1    & 0.1   & 1.0   & 0.3   \\
   &    & 0.2    &    &    &    & $\highest{98.8}$   &    & 0.2   & 0.9   \\
   &    & 21.5   &    & 2.4   &    &     & $\highest{75.5}$  & 0.2   & 0.3   \\
   &    & 3.0    &    & 1.0   &    & 0.3    &    & $\highest{94.9}$  & 0.7   \\
   &    & 0.6    &    &    &    & 0.2    &    &    & $\highest{99.1}$  \\
\bottomline
\multicolumn{3}{c}{$\textcolor{red}{(a,b,r): 96.87\%}$} \\ \topline
$\highest{97.0}$  & 0.1   & 0.9    & 0.5   & 0.5   & 1.0   &     &    & 0.1   &    \\
3.5   & $\highest{90.4 }$ &     & 4.4   & 0.9   & 0.7   &     &    &    &    \\
0.1   &    & $\highest{98.5}$   &    & 0.1   & 0.1   &     & 0.2   & 0.1   & 0.9   \\
1.6   & 0.2   & 0.2    & $\highest{96.9}$  & 0.2   & 0.7   & 0.2    &    &    &    \\
0.1   &    & 0.3    & 0.3   & $\highest{97.3}$  & 1.3   &     & 0.2   & 0.4   & 0.1   \\\midrule
0.2   &    &     &    &    & $\highest{99.4}$  & 0.1    &    & 0.9   & 0.1   \\
0.2   &    &     &    &    &    & $\highest{99.0}$   &    & 0.2   & 0.7   \\
0.2   &    & 7.9    &    & 0.3   &    & 0.3    & $\highest{86.7}$  & 3.3   & 1.2   \\
   &    &     &    & 0.2   & 0.2   & 2.3    &    & $\highest{97.4}$  &    \\
0.3   &    & 0.6    &    & 0.4   & 0.1   & 5.8    &    & 0.1   & $\highest{92.8}$  \\
\bottomline
\multicolumn{3}{c}{$\textcolor{red}{(r, \theta): 93.51\%}$}  \\ \topline
$\highest{91.7}$  & 0.2   & 1.8    &    & 4.4   & 0.5   & 0.1    & 1.2   & 0.2   &    \\
5.1   & $\highest{86.2}$  & 0.2    & 0.7   & 7.7   &    &     &    &    &    \\
0.2   &    & $\highest{96.8}$   &    & 0.5   &    &     & 1.9   & 0.3   & 0.3   \\
9.5   & 13.5  &     & $\highest{56.1}$  & 16.9  & 1.8   & 0.7    & 1.3   & 0.2   &    \\
0.1   & 0.1   & 1.3    &    & $\highest{96.6}$  & 0.1   &     & 1.1   & 0.6   & 0.1   \\\midrule
0.1   & 0.1   & 0.6    &    & 3.3   & $\highest{94.1}$  & 0.1    &    & 1.3   & 0.4   \\
   &    &     &    &    &    & 99.7   &    & 0.2   & 0.2 \\
   &    & 11.0   &    & 0.3   &    &     & $\highest{86.0 }$ & 2.4   & 0.2   \\
   &    &     &    &    &    & 1.0    & 0.2   & $\highest{98.6}$  & 0.2   \\
   &    & 6.7    &    & 0.1   & 0.1   & 1.9    & 0.6   & 1.4   & $\highest{89.2}$  \\
\bottomline
\multicolumn{3}{c}{$\textcolor{red}{\mathbf{z}: 98.16\%}$} \\ \topline
$\highest{97.8}$  & 0.1   & 1.9    &    & 0.2   & 0.1   &     &    &    &    \\
1.4   & $\highest{97.4}$  & 0.2    & 0.7   & 0.2   &    &     &    &    &    \\
0.4   &    & $\highest{99.0}$   &    & 0.1   & 0.1   &     &    &    & 0.4   \\
4.2   & 1.8   & 1.1    & $\highest{90.2}$  & 1.6   & 1.1   &     &    &    &    \\
   & 0.2   & 1.8    &    & $\highest{96.4}$  & 1.0   &     &    &    & 0.6   \\\midrule
   &    & 0.4    &    & 0.1   & $\highest{98.9}$  & 0.1    &    &    & 0.5   \\
   &    &     &    &    &    & 10  &    &    &    \\
   &    & 4.9    &    &    &    & 0.2    & $\highest{94.4}$  & 0.5   &    \\
   &    & 1.2    &    &    &    & 0.5    &    & $\highest{98.3}$  &    \\
   &    & 0.9    &    & 0.1   & 0.1   &     &    &    & $\highest{98.9}$  \\
\bottomline
\end{tabular}
}
\caption{Confusion matrices for 4 real-valued  baselines and our complex-valued CNN.  The method and the overall accuracy is listed at the top left corner of each table. The order of categories is the same as that in Fig. \ref{fig0}.}
\label{tab1}
\end{table}

This performance gain has to come from the group equivariant property of our convolution and the group invariant property of our CNN classifier.  The group that acts on the complex numbers is $\mathbf{R}\setminus \left\{0\right\}\times \textsf{SO}(2)$.  Our equivariance and invariance properties guarantee that our learned CNN is invariant to scaling and planar rotations, unlike any standard real-valued CNN architecture. 
Table \eqref{tab1} also suggests that our learned CNN is more robust to the imbalanced training data.  For example, on the smallest class  `BTR70' with test set size $429$, our model correctly classifies $406$ samples while the baseline correctly classifies only $172$ samples.  

Among the real-valued baselines,  just the magnitude $r$ alone gives a better classification accuracy than the two-channel real-valued representation $(a,b)$.  Their combination $(a,b,r)$ achieves a classification accuracy of $96.87\%$,  with $2\%$ improvement over the  magnitude only representation of $r$. 
The polar representation $(r, \theta)$ is better than the two-channel real-imaginary representation $(a,b)$, but is in fact worse than the magnitude $r$ only representation. A natural question is whether phase information is useful at all.

\begin{table}[htbp]
\centering
\scalebox{0.72}{
\begin{tabular}{|c|c|c|c|c|c|c|c|c|c|}
\multicolumn{3}{c}{ $\textcolor{red}{(a,b): 45.98\%}$ }\\
\topline
   &    & $\highest{100}$  &    &    &    &     &    &    &    \\
   &    & $\highest{100}$  &    &    &    &     &    &    &    \\
   &    & $\highest{100}$  &    &    &    &     &    &    &    \\
   &    & $\highest{100}$  &    &    &    &     &    &    &    \\
   &    & $\highest{100}$  &    &    &    &     &    &    &    \\\midrule
   &    & $\highest{100}$  &    &    &    &     &    &    &    \\
   &    & $\highest{100}$  &    &    &    &     &    &    &    \\
   &    & $\highest{100}$  &    &    &    &     &    &    &    \\
   &    & $\highest{100}$  &    &    &    &     &    &    &    \\
   &    & $\highest{100}$  &    &    &    &     &    &    &    \\
\bottomline

\multicolumn{3}{c}{$\textcolor{red}{\mathbf{z:} 97.00\%}$ } \\
\topline

$\highest{95.6}$  & 0.3   & 2.9    & 0.8   & 0.2   & 0.2   &     &    &    &    \\
2.6   & $\highest{94.4}$  & 0.2    & 2.3   & 0.5   &    &     &    &    &    \\
0.6   &    & $\highest{97.9}$   &    & 0.4   & 0.3   &     &    & 0.2   & 0.4   \\
2.9   & 2.0   & 1.6    & $\highest{90.7}$  & 2.0   & 0.9   &     &    &    &    \\
   & 0.3   & 1.5    & 0.3   & $\highest{94.8}$  & 2.6   &     & 0.1   & 0.3   & 0.3   \\\midrule
   &    & 0.6    &    & 0.5   & $\highest{98.4}$  & 0.1    &    & 0.1   & 0.4   \\
   &    &     &    &    &    & $\highest{99.7 }$  &    & 0.3   & 1.7   \\
   &    & 6.8    &    & 0.7   & 0.2   &     & $\highest{91.1}$  & 0.9   & 0.3   \\
   & 0.2   & 0.7    &    &    &    & 0.3    &    & $\highest{98.8}$  &    \\
   &    & 1.5    &    & 0.4   & 0.3   & 0.1    & 0.1   & 0.1   & $\highest{97.6 }$\\
\bottomline
\end{tabular}
}
\caption{Confusion matrices for the baseline model $(a,b)$ (top) and our model (bottom) applied to normalized complex numbers.  Same convention as Table \ref{tab1}. With an overall accuracy of 97\% over  the baseline accuracy 46\%, our complex-valued CNN brings significant discrimination power out of the phase information alone.}
\label{tab2}
\end{table}

\begin{figure}[!htp]
\centering
\includegraphics[width=0.49\textwidth,clip]{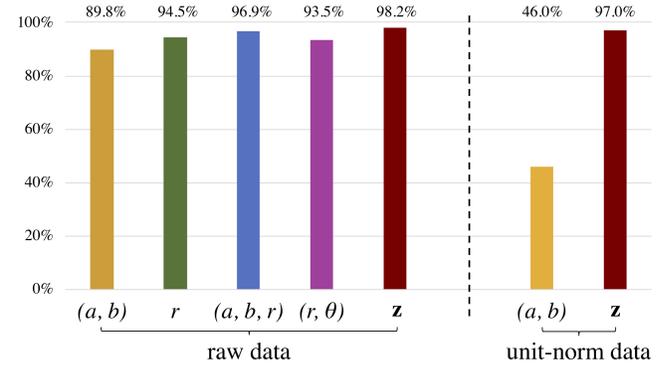}
\caption{MSTAR classification accuracies by real-valued baseline CNNs and our complex-valued CNN, with raw and normalized complex number inputs.
}\label{fig0.25}
\end{figure}

\noindent
{\bf How useful is phase alone?}  We remove any useful information in the magnitude by normalizing each complex number to norm $1$.  On the normalized complex numbers, Table \eqref{tab2} shows the classification confusion matrix for the baseline $(a,b)$ CNN model and our model.  The real-valued CNN achieves an overall accuracy of $45.98\%$, 
with all the test set classified as the largest class which consists of $45.98\%$ samples of the entire dataset.  That is, the real-valued CNN is completely confused by the phase and unable to tease apart different classes.  On the other hand, our model gives a surprisingly high accuracy of  $97\%$, only $1\%$ less than our result on the raw complex numbers which contains the class-discriminative magnitude.  

Fig. \eqref{fig0.25} compares the classification accuracies in different settings.
The stark contrast in real- and complex-valued CNNs to phase data alone demonstrates not only the effectiveness of our complex-valued CNN due to its invariance to $G$, but also the richness of the phase information alone.

\begin{table}[tp]
\centering
\begin{tabular}{c|c|r}
\toprule
\bf CNN model & 
\bf domain representation & 
\bf \# parameters  \\ \midrule
real & $(a,b)$ & $530,170$ \\ \hline
real & $r$ & $530,026$ \\ \hline
real & $(a,b,r)$ & $530,314$ \\ \hline
real & $(r, \theta)$ & $530,170$ \\ \hline
complex & $\mathbf{z}$ & $\mathbf{44,826}$ \\
\bottomrule
\end{tabular}
\caption{CNN model size comparison.  Our complex-valued CNN is $8\%$ of the baseline real-valued CNN model size.}
\label{tab5}
\end{table} 
 
 \begin{figure*}[!tp]
\centering
\includegraphics[height=0.335\textheight,clip]{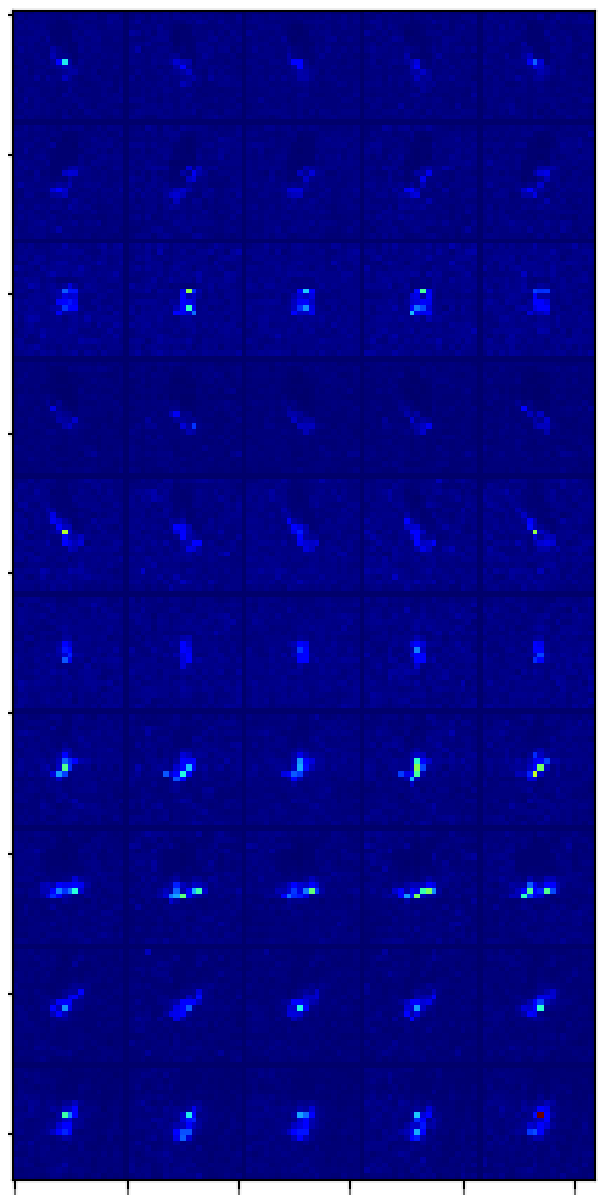}%
\includegraphics[height=0.335\textheight,clip]{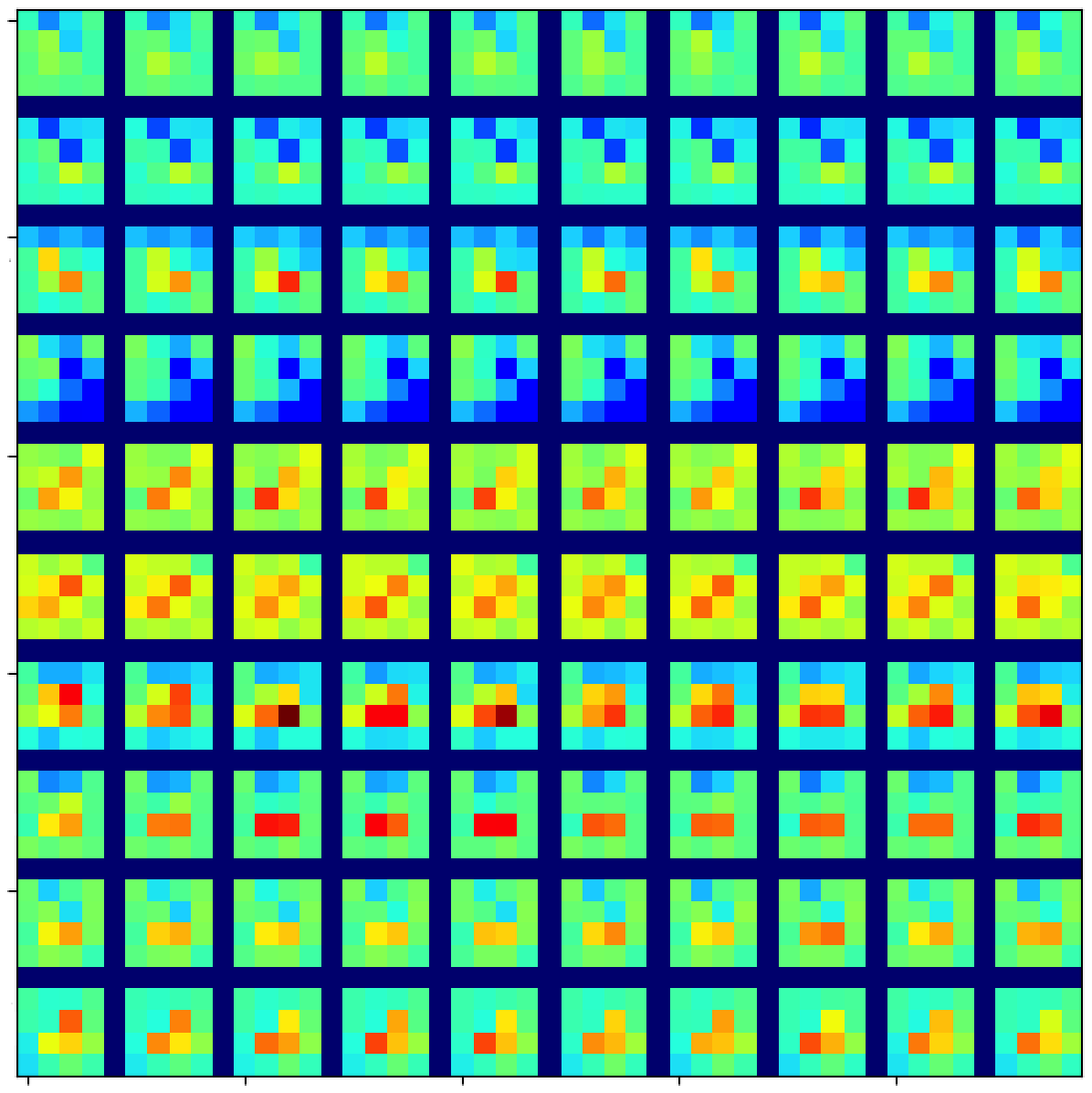}%
\includegraphics[height=0.335\textheight,clip]{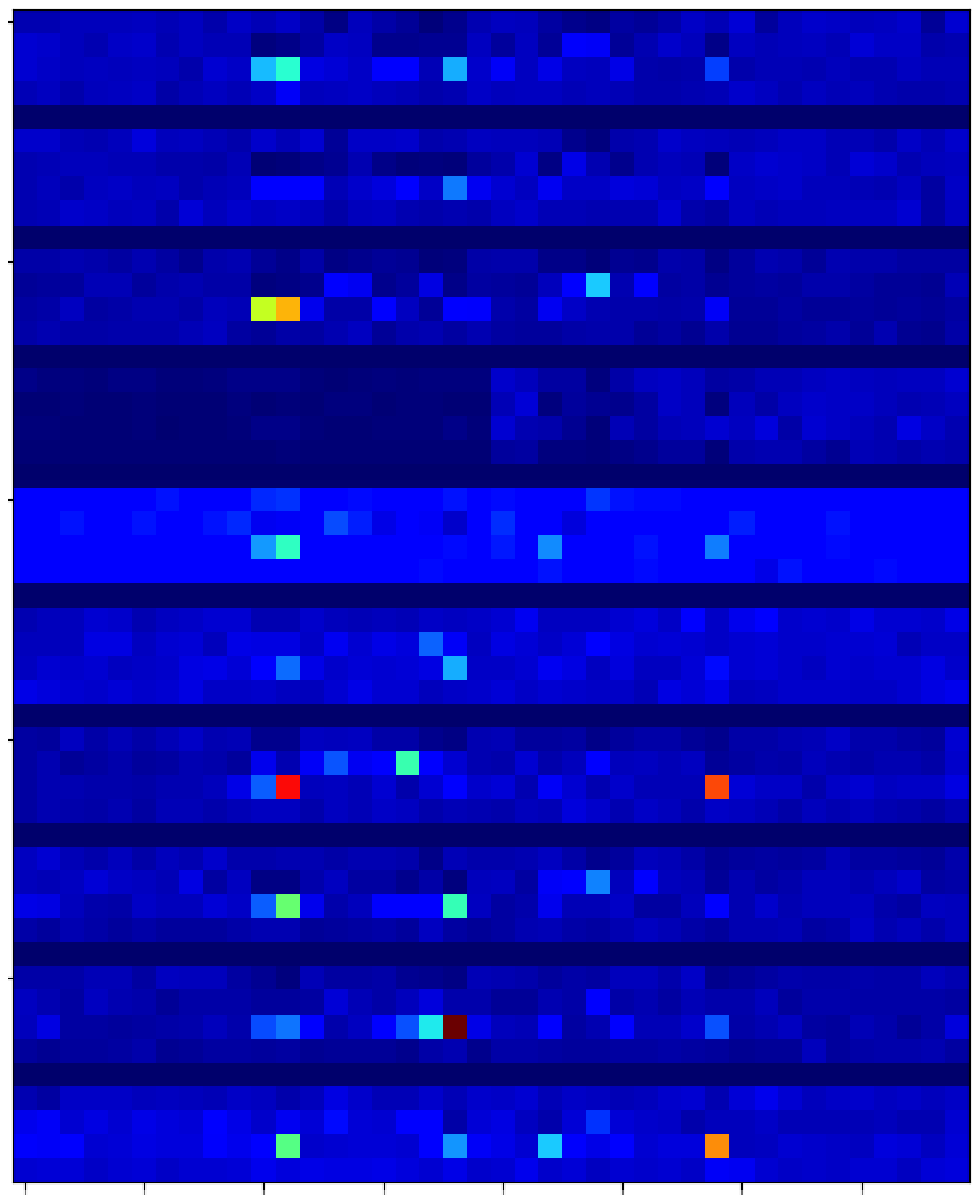}
\caption{Sample MSTAR filter responses of our model after the first, second and third conv layer.  Each row corresponds to the same image; each column represents a particular channel's absolute valued response. }
\label{fig2}
\end{figure*}

\noindent
{\bf Our complex-valued CNN is better and leaner.} 
Table \eqref{tab5} lists the total number of 
parameters used in each CNN model. 
As our complex-valued CNN captures the natural equivariance and invariance in the non-Euclidean complex number range space,  which standard CNNs fail to do,
our model achieves a higher accuracy with a significant (more than 90\%) parameter reduction.  

\noindent
{\bf CNN visualization.}
Fig. \eqref{fig2} shows examples of filter responses at three convolution layers on the representative images in Fig. \eqref{fig0}.  The first convolution layer produces basically  blurred versions of the input image. 
From the second convolution layer onward, the filter response patterns grow more divergent for different classes. While we show one sample output from each class, the patterns within each class are similar.  For classes `D7', `T62', `ZIL131', the filter responses are higher than the other classes. Furthermore, the last convolution layer shows significantly different patterns between different classes.

 \subsection{RadioML Experiments}

\begin{figure}[!bp]
\centering
\includegraphics[width=0.49\textwidth,clip]{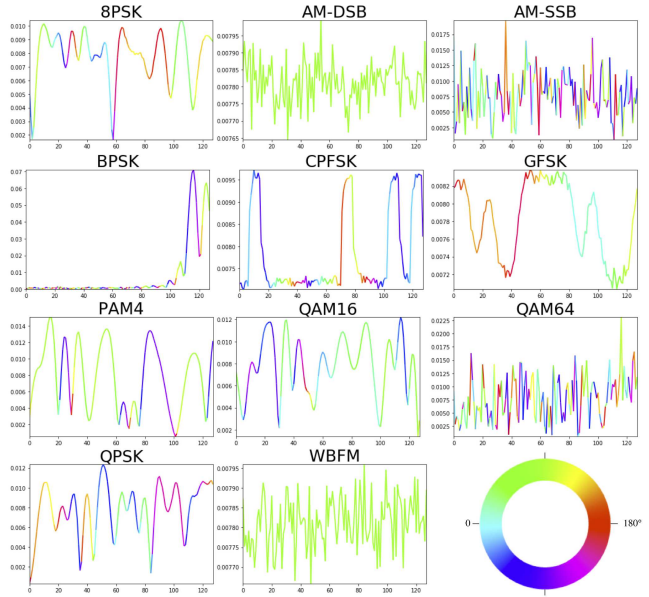}
\caption{RadioML data samples. We plot one sample per class at SNR 18. We use the HSV colormap to encode and visualize the phase of complex valued 1D signals.}\label{radioml}
\end{figure}

\begin{figure*}[tp]
\centering
\includegraphics[height=0.28\textheight,clip]{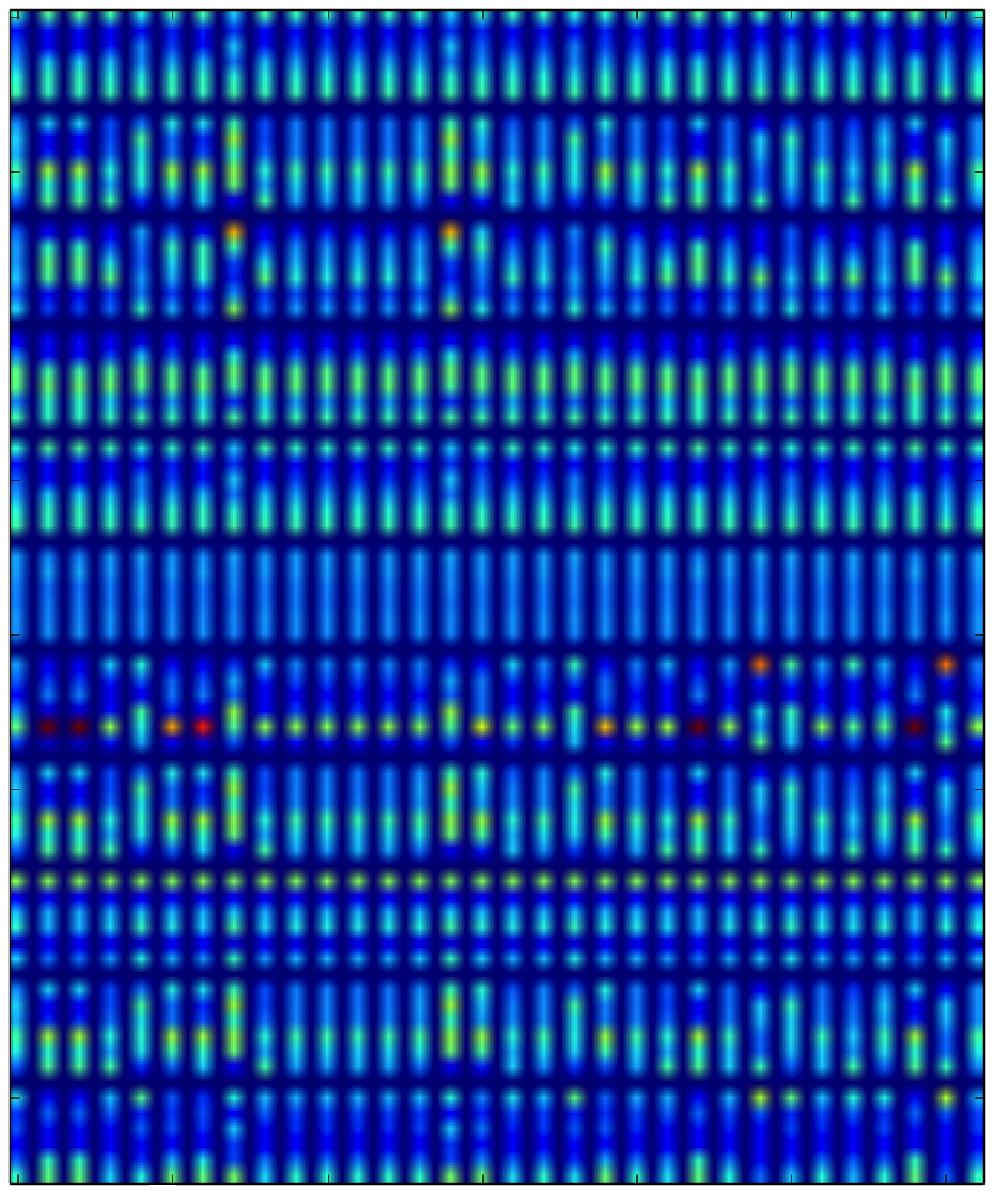}%
\includegraphics[height=0.28\textheight,clip]{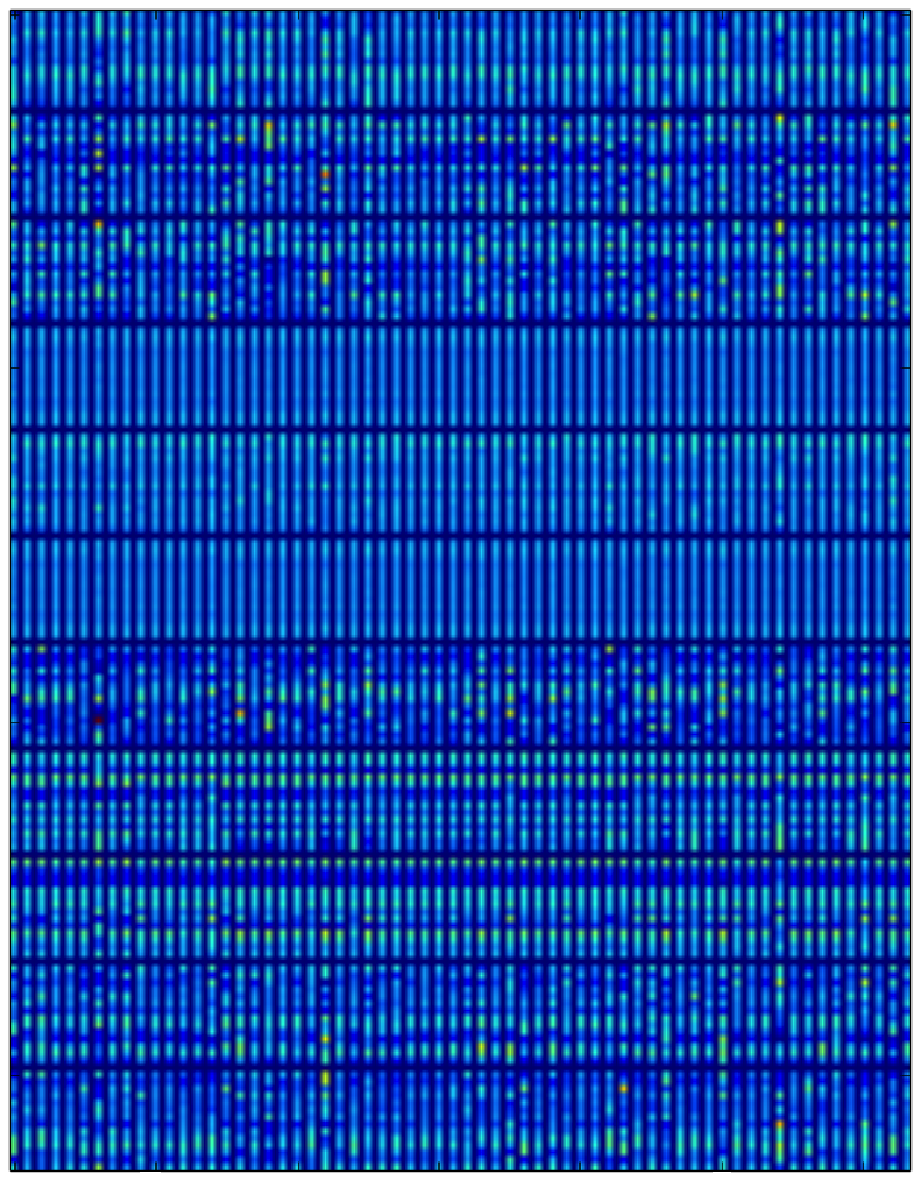}%
\includegraphics[height=0.28\textheight,clip]{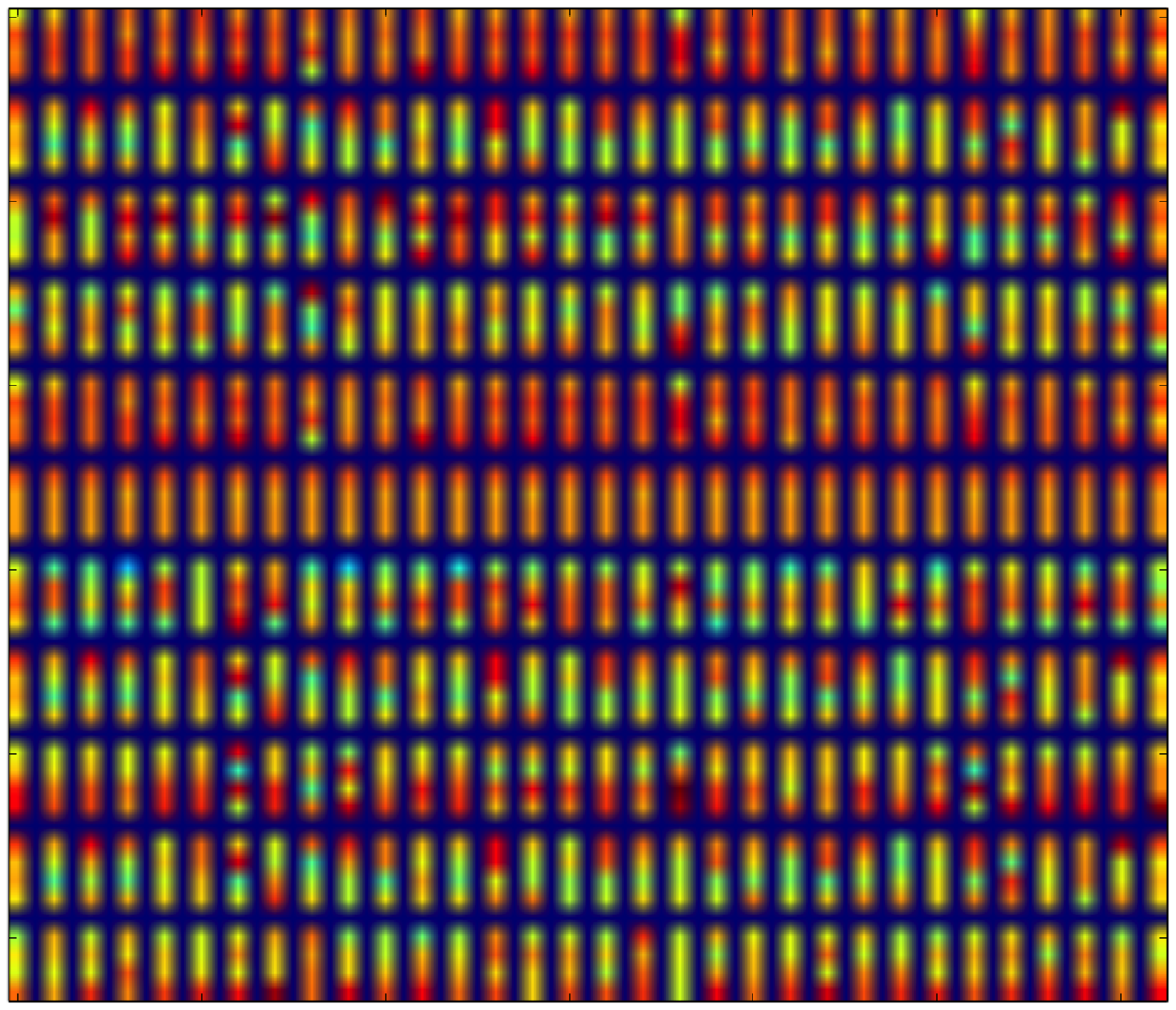}
\caption{Representative filter outputs after the first, second, third convolutional layers (absolute valued responses) of our complex-valued network on the RadioML data.  Same convention as Fig. \ref{fig2}.}\label{fig1}
\end{figure*}

\begin{figure}[bp]
\centering
\includegraphics[width=0.49\textwidth]{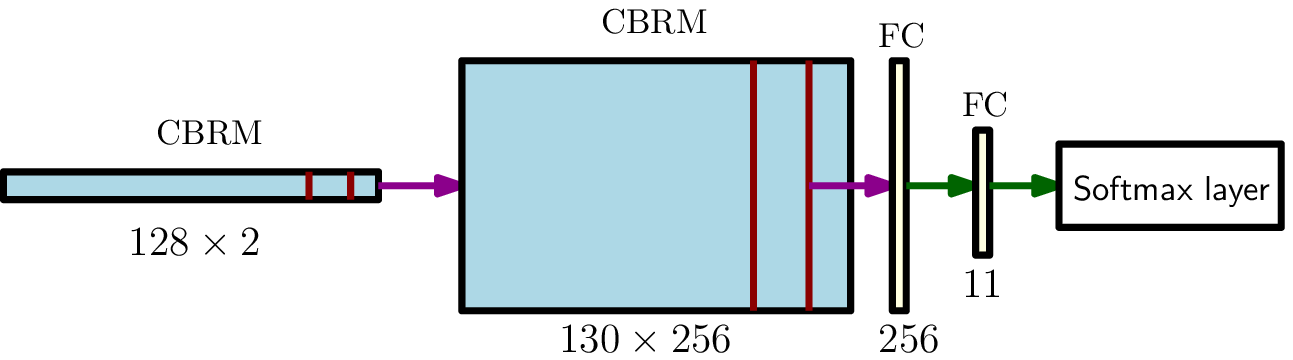}\\
\includegraphics[width=0.49\textwidth]{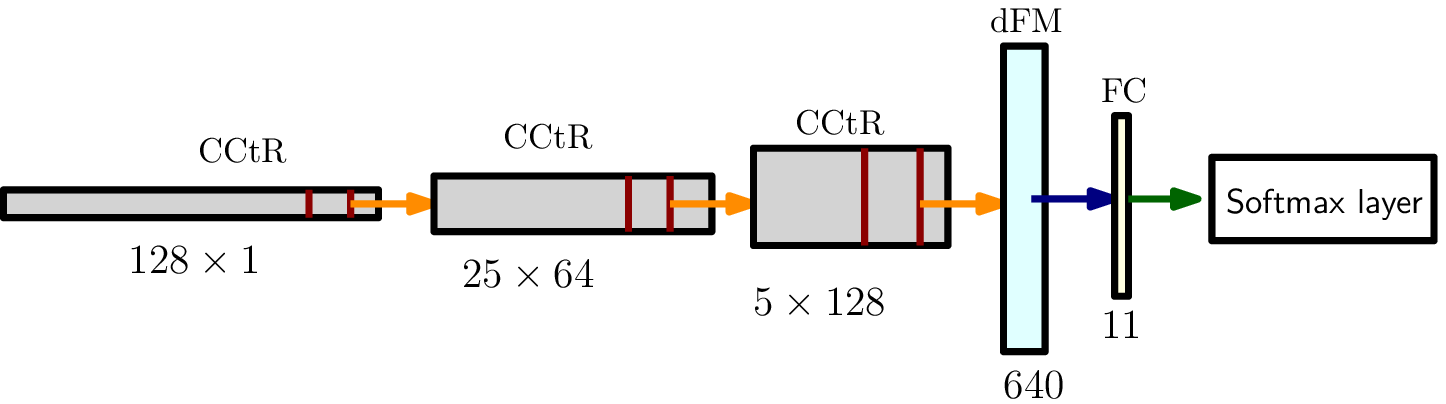}
\caption{Real-valued CNN  model and our complex-valued model for RadioML.  CBRM denotes Convolution, Batch-Normalization, ReLU,  and Pooling. CCtR denotes our Complex-valued Convolution and tangent ReLU, and dFM our distance transform with respect to the Fr{\'e}chet mean.
}\label{fig1.1}
\end{figure}

\noindent
{\bf RadioML dataset.}  RF modulation operates on both discrete binary alphabets (digital modulations) and continuous alphabets (analog modulations).  Over each modem the known data is modulated and then exposed to channel effects using GNU Radio.  It is then segmented into short-time windows in a fashion similar to how
a continuous acoustic voice signal is typically windowed for voice recognition
tasks.  Fig. \eqref{radioml} visualizes these 1D complex-valued time series as colored lines.  There are $220,000$ samples in RadioML \cite{convnetmodrec,rml_datasets}. 
We use a 50-50 train-test split and 10 random runs as in our MSTAR experiments.

\noindent
{\bf RadioML baseline.}
It consists of two convolutional and two fully connected layers as used in \cite{convnetmodrec}.  The convolution  kernel is of size $3$ with $256$ and $80$ channels respectively. Each convolutional layer is followed by ReLU and dropout layers.  This network has $2,830,491$ parameters.

\noindent
{\bf Our RadioML CNN model.}
It has two complex convolutional layers of stride 5, kernel sizes 7 and 5, the numbers of channels 64 and 128, followed by an invariant distance transform layer and a final softmax layer for classification. 
Fig. \ref{fig1.1} shows both the real-valued baseline CNN and our complex-valued CNN architectures.  We use ADAM optimizer \cite{kingma2014adam} with learning rate 0.05 and mini-batch size 500.

Our complex-valued CNN has only $299,117$ parameters, i.e., roughly $10\%$ of the baseline model, yet it can achieve test accuracy $70.23\%$, on par with $70.68\%$ of the baseline real-valued CNN model.  This lean model result is consistent with our MSTAR experiments.  Fig. \eqref{fig1} also shows that discriminative filter response patterns emerge quickly from various smoothing effects of convolutional layers.

\section{Summary} 
\label{sec4}
We take a manifold view on complex-valued data and  present a novel CNN theory.
Our convolution from Fr{\'e}chet mean filtering is equivariant and our distance transform is invariant to complex-valued scaling, an inherent ambiguity in the complex value range space.  

Our experiments on  MSTAR and RadioML  demonstrate that our complex-valued CNN classifiers can deliver better accuracies with a surreal leaner CNN model, at a fraction of the real-valued CNN model size.

By representing a complex number as a point on a manifold instead of two independent real-valued data points, our model is more robust to imbalanced classification and far more powerful at discovering discriminative information in the phase data alone.

\noindent{\bf Acknowledgements.}
This research was supported, in part, by Berkeley Deep Drive and DARPA.  The views, opinions and/or findings expressed are those of the author and should not be interpreted as representing the official views or policies of the Department of Defense or the U.S. Government.

\bibliographystyle{IEEEbib}
\bibliography{references}

\end{document}